\documentclass[10pt,twocolumn,letterpaper]{article}

\usepackage{cvpr}
\usepackage{times}
\usepackage{epsfig}
\usepackage{graphicx}
\usepackage{amsmath}
\usepackage{amssymb}
\usepackage{amsthm}
\usepackage{subcaption}
\usepackage{bm}
\usepackage{gensymb}
\usepackage{booktabs}
\usepackage{multirow}
\usepackage{caption}
\usepackage{enumitem}

\usepackage[toc,page]{appendix}
\usepackage{array}
\newcolumntype{C}[1]{>{\centering\let\newline\\\arraybackslash\hspace{0pt}}m{#1}}
\newcolumntype{?}{!{\vrule width 1pt}}

\usepackage[breaklinks=true,bookmarks=false]{hyperref}

\cvprfinalcopy 

\def\cvprPaperID{5882} 

\ifcvprfinal\fi
\begin{document}

\title{Multiview 2D/3D Rigid Registration via\\ a Point-Of-Interest Network for Tracking and Triangulation}

\author{Haofu Liao\textsuperscript{1}*, Wei-An Lin\textsuperscript{2}*, Jiarui Zhang\textsuperscript{3}, Jingdan Zhang\textsuperscript{4}, Jiebo Luo\textsuperscript{1}, S. Kevin Zhou\textsuperscript{5,6}\\
\textsuperscript{1}University of Rochester \hspace{2em} \textsuperscript{2}University of Maryland, College Park \hspace{2em} \textsuperscript{3}Rutgers University\\
 \textsuperscript{4}Z2W Corporation \hspace{2em} \textsuperscript{5}Chinese Academy of Sciences \hspace{2em} \textsuperscript{6}Peng Cheng Laboratory, Shenzhen
}

\maketitle

\begin{abstract}

We propose to tackle the multiview 2D/3D rigid registration problem via a Point-Of-Interest Network for Tracking and Triangulation (POINT$^2$). POINT$^2$ learns to establish 2D point-to-point correspondences between the pre- and intra-intervention images by tracking a set of point-of-interests (POIs). The 3D pose of the pre-intervention volume is then estimated through a triangulation layer. In POINT$^2$, the unified framework of the POI tracker and the triangulation layer enables learning informative 2D features and estimating 3D pose jointly. In contrast to existing approaches, POINT$^2$ only requires a single forward-pass to achieve a reliable 2D/3D registration. As the POI tracker is shift-invariant, POINT$^2$ is more robust to the initial pose of the 3D pre-intervention image. Extensive experiments on a large-scale clinical cone-beam computed tomography dataset show that the proposed POINT$^2$ method outperforms the existing learning-based method in terms of accuracy, robustness and running time. Furthermore, when used as an initial pose estimator, our method also improves the robustness and speed of the state-of-the-art optimization-based approaches by ten folds.
{\let\thefootnote\relax\footnote{{* indicates equal contributions.}}}
\setcounter{footnote}{0}

\end{abstract}

\section{Introduction} \label{sec:intro}

\begin{figure}[t!]
\centering
\begin{subfigure}{0.5\linewidth}
  \centering
  \includegraphics[width=\linewidth]{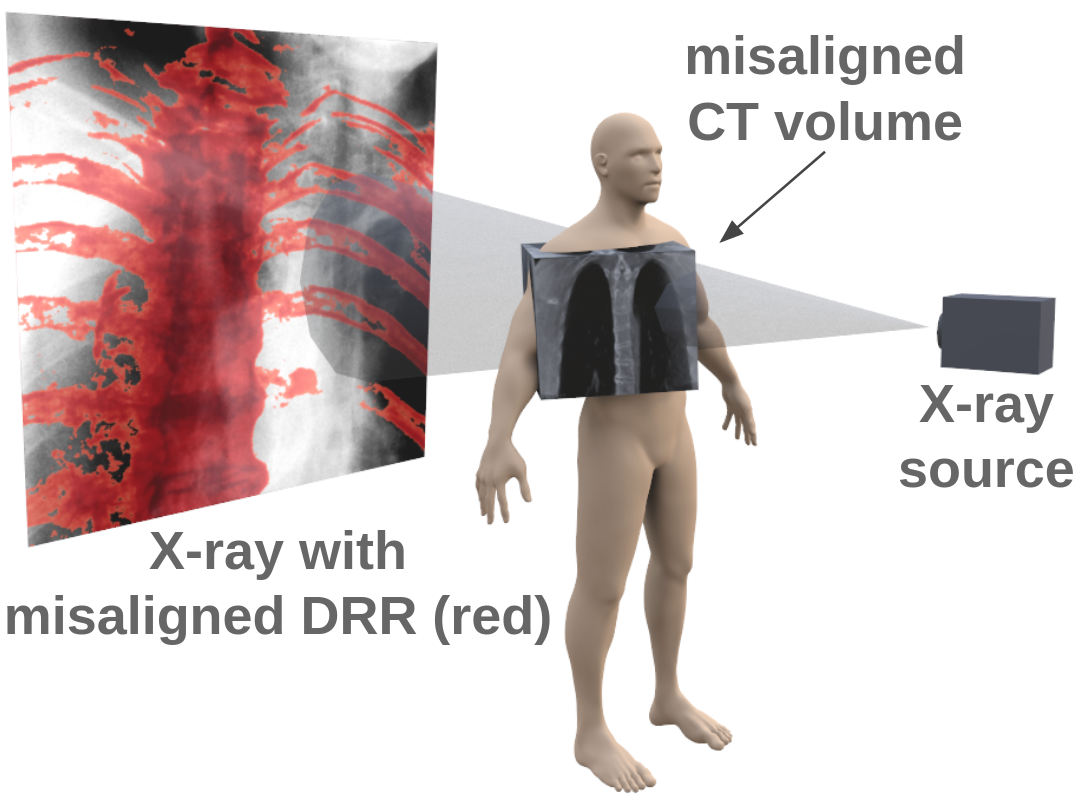}
  \caption{before registration}
\end{subfigure}%
\begin{subfigure}{0.5\linewidth}
  \centering
  \includegraphics[width=\linewidth]{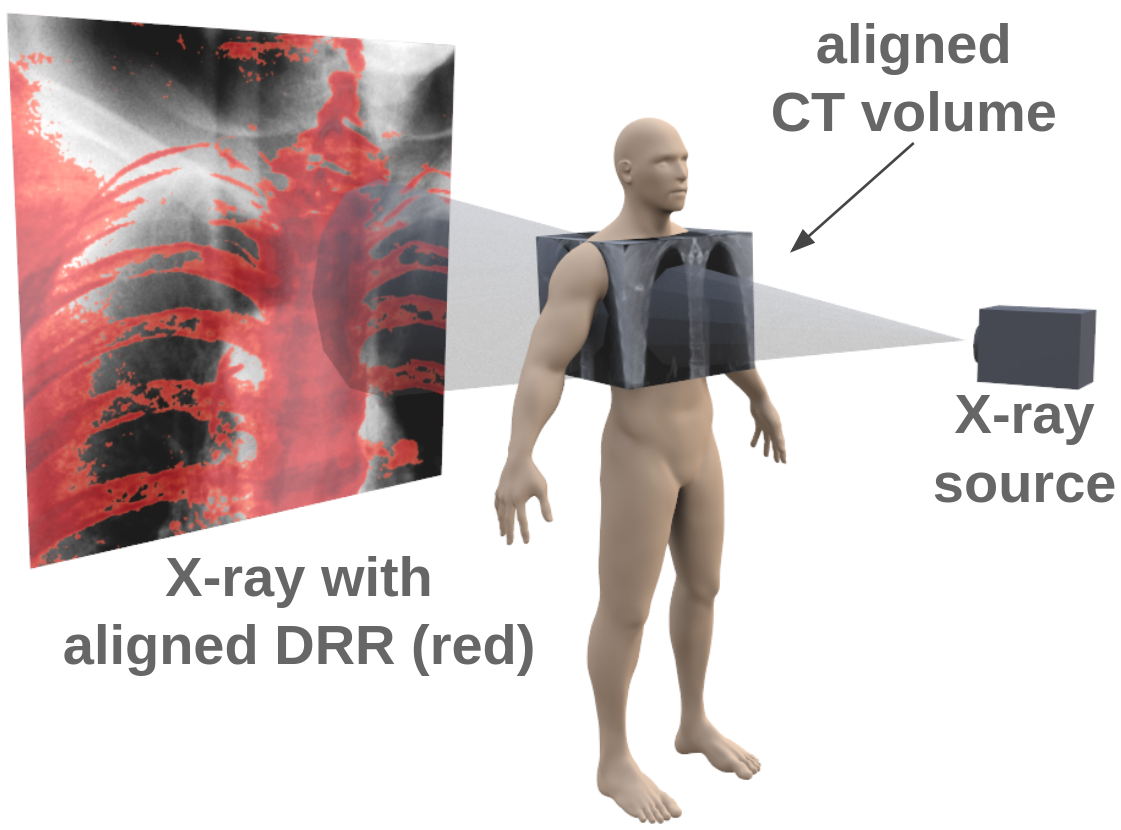}
  \caption{after registration}
\end{subfigure}
\caption{Overlay of the DRRs and X-rays before and after the 2D/3D registration. For visualization purpose, only the bone region of the DRRs are projected and recolored with red to distinguish from the X-rays.}
\label{fig:alignment}
\end{figure}

In 2D/3D rigid registration for intervention, the goal is to find a rigid pose of a pre-intervention 3D data, e.g., computed tomography (CT), such that it aligns with a 2D intra-intervention image of a patient, e.g., fluoroscopy. In practice, CT is usually a preferred 3D pre-intervention data as digitally reconstructed radiographs (DRRs) can be produced from CT using ray casting~\cite{sherouse1990computation}. The generation of DRRs simulates how an X-ray is captured, which makes them visually similar to the X-rays. Therefore, they are leveraged to facilitate the 2D/3D registration as we can observe the misalignment between the CT and patient by directly comparing the intra-intervention X-ray and the generated DRR (See Figure \ref{fig:alignment} and Section \ref{sec:problem} for details).

One of the most commonly used 2D/3D registration strategies~\cite{markelj2012review} is through an optimization-based approach, where a similarity metric is first designed to measure the closeness between the DRRs and the 2D data, and then the 3D pose is iteratively searched and optimized for the best similarity score. However, the iterative pose searching scheme usually suffers from two problems. First, the generation of DRRs incurs high computation, and the iterative pose searching requires a significant number of DRRs for the similarity measure, making it computationally slow. Second, iterative pose searching relies on a good initialization. When the initial position is not close enough to the correct one, the method may converge to local extrema, and the registration fails. Although many studies have been proposed to address these two problems~\cite{ de20163d,otake2013robust,otake2012intraoperative,khamene2006automatic,dey2006targeted,jans20063d,russakoff2005fast}, trade-offs still have to be made between sampling good starting points and less costly registration.

In recent years, the development of deep neural networks (DNNs) has enabled a learning-based strategy for medical image registration~\cite{DBLP:conf/aaai/MiaoPFTMML18,toth20183d,DBLP:conf/aaai/LiaoMTGKMC17,miao2016cnn} that aims to estimate the pose of the 3D data without searching and sampling the pose space at a large scale. Despite the efficiency, there are still two limitations of the existing learning-based methods. First, the learning-based methods usually require generating a huge number of DRRs for training. The corresponding poses for the DRRs have to be dense in the entire searching space to avoid overfitting. Considering that the number of required DRRs is exponential with respect to the dimension of the pose space (which is usually six), this is computationally prohibitive, thus making the learning-based methods less reliable during testing. Second, the current state-of-the-art learning-based methods~\cite{DBLP:conf/aaai/MiaoPFTMML18,toth20183d,DBLP:conf/aaai/LiaoMTGKMC17}  require an iterative refinement of the estimated pose and use DNNs to predict the most plausible update direction for faster convergence. However, the iterative approach still introduces a non-negligible computational cost, and the DNNs may direct the searching to an unseen state, which fails the registration quickly.

In this paper, we introduce a novel learning-based approach, which is referred to as a Point-Of-Interest Network for Tracking and Triangulation (POINT$^2$). POINT$^2$ directly aligns the 3D data with the patient by using DNNs to establish a point-to-point correspondence between multiple views\footnote{A different view indicates the DRR or X-ray is captured at a different projection angle.} of DRRs and X-ray images. The 3D pose is then estimated by aligning the matched points. Specifically, these are achieved by tracking a set of points of interest (POIs). For 2D correspondence, we use the POI tracking network to map the 2D POIs from the DRRs to the X-ray images. For 3D correspondence, we develop a triangulation layer that projects the tracked POIs in the X-ray images of multiple views back into 3D. We highlight that since the point-to-point correspondence is established in a shift-invariant manner, the requirement of dense sampling in the entire pose space is avoided.

The contributions of this paper are as follows:
\begin{itemize}[noitemsep,topsep=0pt]
\item A novel learning-based multiview 2D/3D rigid registration method that directly measures the 3D misalignment by exploiting the point-to-point correspondence between the X-rays and DRRs, which avoids the costly and unreliable iterative pose searching, and thus delivers faster and more robust registration.

\item A novel POI tracking network constructed using a Siamese U-Net with POI convolution to enable a fine-grained feature extraction and effective POI similarity measure, and more importantly, to offer a shift-invariant 2D misalignment measure that is robust to in-plane offsets\footnote{In-plane/out-plane offset refers to the translation and rotation offset within/outside the DRR or X-ray images.}.

\item A unified framework of the POI tracker and the triangulation layer, which enables (i) end-to-end learning of informative 2D features and (ii) 3D pose estimation.

\item An extensive evaluation on a large-scale and challenging clinical cone-beam CT (CBCT) dataset, which shows that the proposed method performs significantly better than the state-of-the-art learning-based approaches, and, when used as an initial pose estimator, it also greatly improves the robustness and speed of the state-of-the-art optimization-based approaches.
\end{itemize}

\section{Related Work}

{\noindent \bf Optimization-Based Approaches.}
Optimization-based approaches usually suffer from high computational cost and is sensitive to the initial estimate. To reduce the computational cost, many works have been proposed to improve the efficiency in hardware-level~\cite{larose2001iterative,khamene2006automatic,otake2012intraoperative} or software-level~\cite{russakoff2005fast,zollei20012d,larose2000transgraph}. Although these works have successfully reduced the DRR generation time to a reasonable range, the overall registration time is still non-negligible~\cite{russakoff2005fast,otake2012intraoperative} and the registration accuracy might be compromised for faster speed~\cite{zollei20012d,russakoff2005fast}. For better initial pose estimation, many attempts have been made by either sampling better initial position~\cite{jans20063d,dey2006targeted}, using multistart strategies~\cite{you2001vivo,otake2013robust}, or a carefully designed objective function that is less sensitive to the initial position selection~\cite{otake2012intraoperative}. However, these methods usually achieve a more robust registration at the cost of longer running time as more locations, and the corresponding DRRs need to be sampled and generated, respectively, to avoid being trapped in the local extrema.

{\vspace{0.5em} \noindent \bf Learning-Based Approaches.} Early learning-based approach~\cite{miao2016cnn} aims to train the DNNs to directly predict the 3D pose given a pair of DRR and X-ray images. However, this approach is generally too ambitious and hence relies on the existence of opaque objects, such as medical implants, that provide strong features for robustness. Alternatively, it has been shown that formulating the registration as a Markov decision process (MDP) is viable~\cite{DBLP:conf/aaai/LiaoMTGKMC17}. Instead of directly regressing the 3D pose, MDP-based methods propose first to train an agent that predicts the most possible search direction and then the registration is iteratively repeated until a fixed number of steps is reached. However, the MDP-based approach requires the agent to be trained on a large number of samples such that the registration can follow the expected trajectory. Though mitigated with a multi-agent design~\cite{DBLP:conf/aaai/MiaoPFTMML18}, it is still inevitable that the neighborhood search may reach an unseen pose and the registration fails. Moreover, the MDP-based approach cannot guarantee convergence and hence limits its registration accuracy. Therefore, the MDP-based approach~\cite{DBLP:conf/aaai/MiaoPFTMML18} is usually used to find a good initial pose for the registration, and a combination with an optimization-based method is applied for better performance. Another possible approach is by directly tracking landmarks from multiple views of X-ray images~\cite{DBLP:conf/miccai/BierUZFAONM18}. However, the landmark-based tracking approach does not make use of the information from the CT volume and requires the landmarks to be present in the X-ray images, making it less robust and applicable to clinical applications.

\section{Methodology}

\begin{figure}[t!]
  \centering
  \includegraphics[height=0.22\textheight]{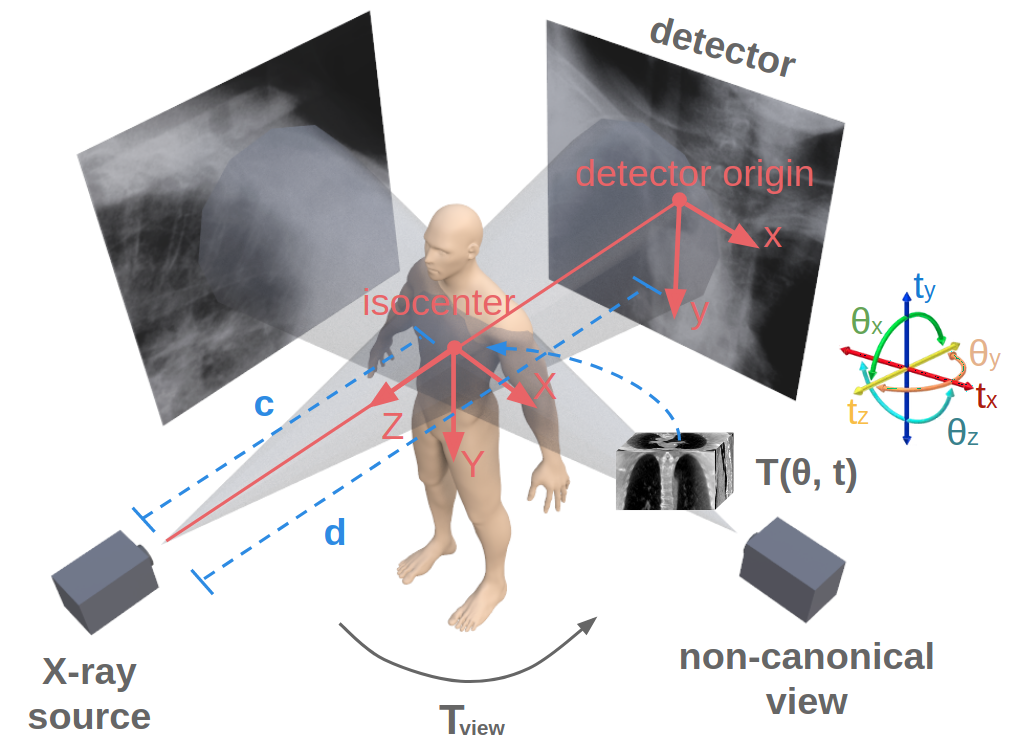}
  \caption{The X-ray imaging model of the canonical-view (bottom-left to upper-right) and a non-canonical view (bottom-right to upper-left).}
\label{fig:camera_model}
\end{figure}

\subsection{Problem Formulation} \label{sec:problem}

Following the convention in the literature~\cite{markelj2012review}, we assume a 2D/3D rigid registration problem and also assume that the 3D data is a CT or CBCT volume, which is the most accessible and allows the generation of DRR. For the 2D data, we use X-rays. As single-view 2D/3D registration is an ill-posed problem (due to the ambiguity introduced by the out-plane offset), X-rays from multiple views are usually captured during the intervention. Therefore, we also follow the literature~\cite{markelj2012review} and tackle a multiview 2D/3D registration problem. Without loss of generality, most of the studies in this work are conducted under two views, and it is easy to extend our work to the cases with more views.

{\vspace{0.5em} \noindent \bf 2D/3D Rigid Registration with DRRs.} In 2D/3D rigid registration, the misalignment between the patient and the CT volume $\mathbf{V}$ is formulated through a transformation matrix $\mathbf{T}$ that brings $\mathbf{V}$ from its initial location to the patient's location under the same coordinate. As illustrated in Figure \ref{fig:camera_model}, $\mathbf{T}$ is usually parameterized by three translations $\mathbf{t}=(t_x, t_y, t_z)^T$ and three rotations $\bm{\theta}=(\theta_x, \theta_y, \theta_z)^T$ about the axes, and can be written as a $4 \times 4$ matrix under the homogeneous coordinate
\begin{equation} \label{eq:trans}
    \mathbf{T} =
    \begin{bmatrix}
        \mathbf{R}(\bm{\theta}) & \mathbf{t} \\
        0 & 1
    \end{bmatrix},
\end{equation}
where $\mathbf{R}$ is the rotation matrix that controls the rotation of $\mathbf{V}$ around the origin.

As demonstrated in Figure \ref{fig:alignment}, casting simulated X-rays through the CT volume creates a DRR on the detector. Similarly, passing a real x-ray beam through the patient's body gives an X-ray image. Hence, the misalignment between the CT volume and the patient can be observed from the detector by comparing the DRR and the X-ray image. Given a transformation matrix $\mathbf{T}$ and a CT volume $\mathbf{V}$, the DRR $\mathbf{I}^{\text{D}}$ can be computed by
\begin{equation} \label{eq:drr}
    \mathbf{I}^{\text{D}}(\mathbf{x}) = \int_{\mathbf{p} \in \bm{l}(\mathbf{x})} \mathbf{V}(\mathbf{T}^{-1} \mathbf{p}) d\mathbf{p},
\end{equation}
where $\bm{l}(\mathbf{x})$, whose parameters are determined by the imaging model, is a line segment connecting the X-ray source and a point $\mathbf{x}$ on the detector. Therefore, let $\mathbf{I}^{\text{X}}$ denote the X-ray image, the 2D/3D registration can be seen as finding the optimal $\mathbf{T}^*$ such that $\mathbf{I}^{\text{X}}$ and $\mathbf{I}^{\text{D}}$ are aligned.

\begin{figure}[t!]
\centering
\begin{subfigure}{0.4\linewidth}
  \centering
  \includegraphics[width=\linewidth]{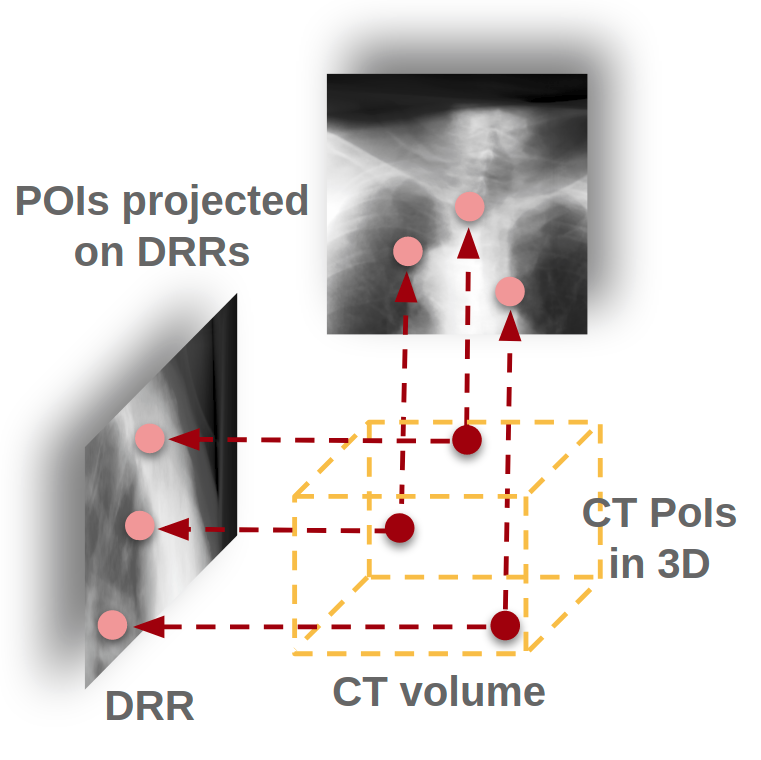}
  \captionsetup{justification=centering}
  \caption{Forward projection \\ from CT to DRR}
\end{subfigure}%
\begin{subfigure}{0.4\linewidth}
  \centering
  \includegraphics[width=\linewidth]{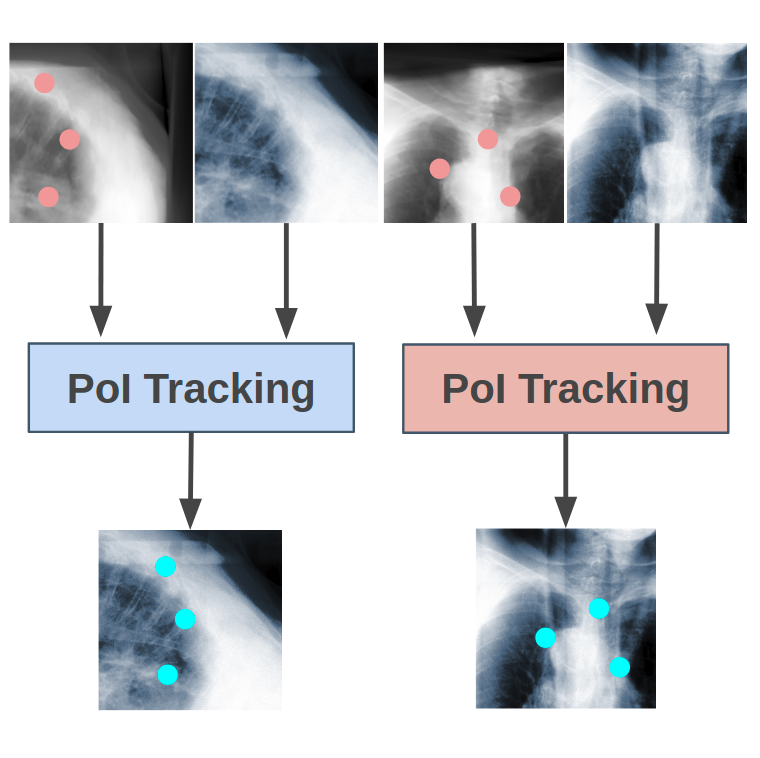}
  \captionsetup{justification=centering}
  \caption{POI tracking \\ from DRR to X-ray}
\end{subfigure}
\begin{subfigure}{0.4\linewidth}
  \centering
  \includegraphics[width=\linewidth]{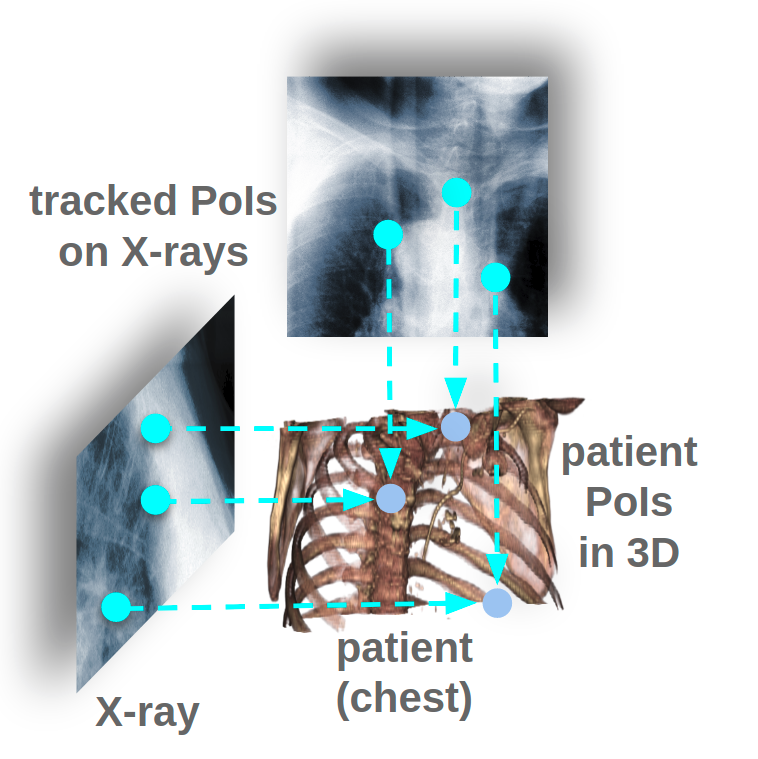}
  \captionsetup{justification=centering}
  \caption{Backward projection \\ from X-ray to patient}
\end{subfigure}%
\begin{subfigure}{0.4\linewidth}
  \centering
  \includegraphics[width=\linewidth]{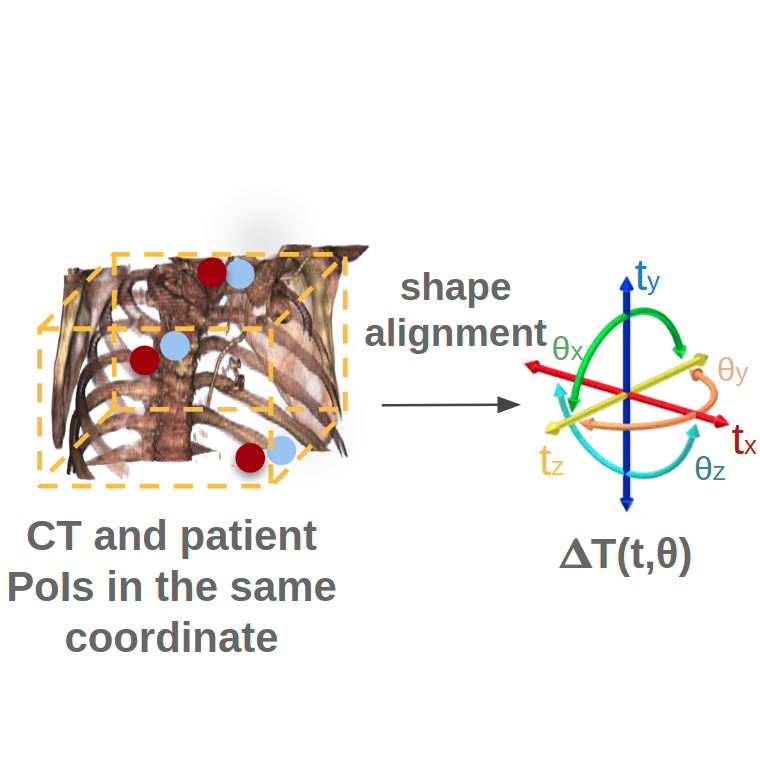}
  \captionsetup{justification=centering}
  \caption{Shape alignment between \\ CT and patient POIs}
\end{subfigure}
\caption{Overview of the proposed POINT$^2$ method. For better visualization, we apply different colormaps to DRR and X-ray images and adjust their contrast.}
\label{fig:overview}
\end{figure}

\begin{figure*}[t!]
  \centering
  \includegraphics[width=0.7\linewidth]{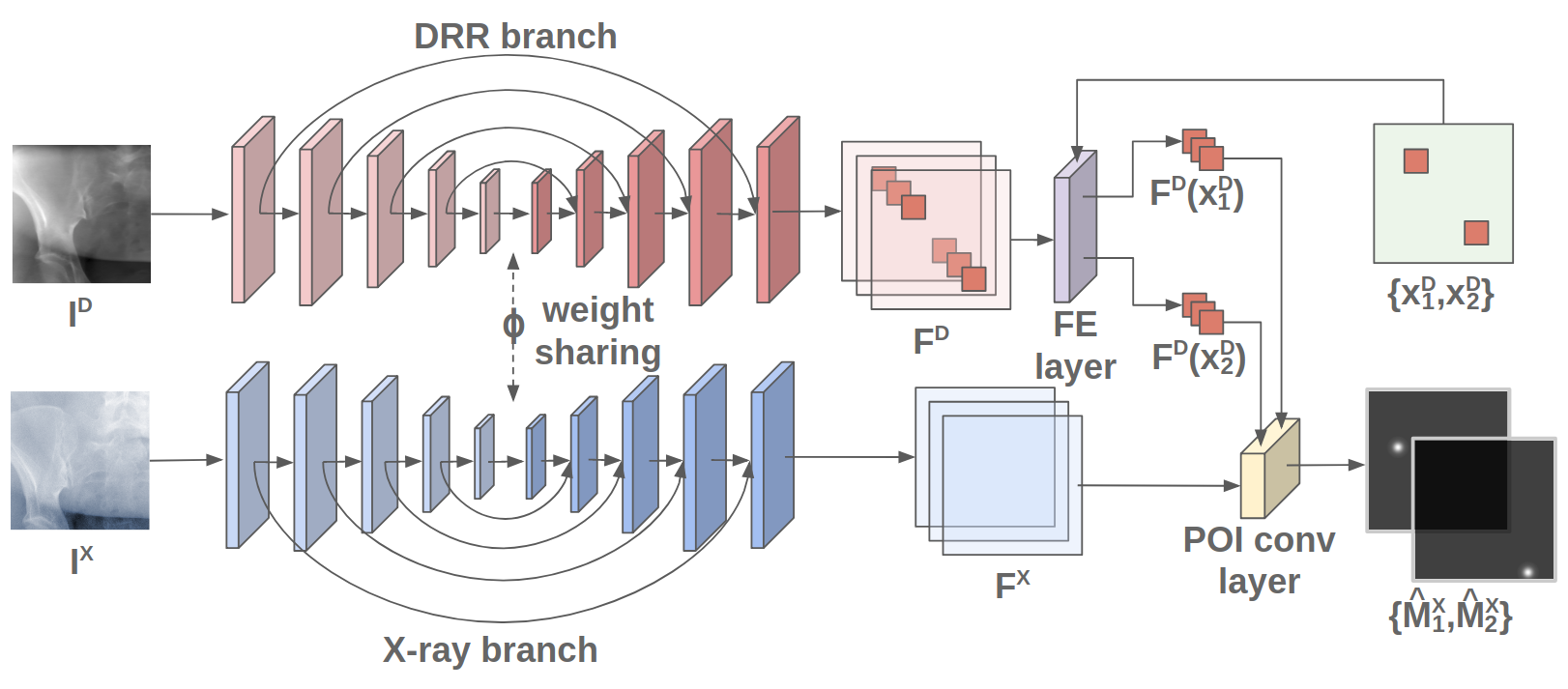}
  \caption{The architecture of the POINT network.}
\label{fig:framework_2d_tracking}
\end{figure*}

{\vspace{0.5em} \noindent \bf X-Ray Imaging Model.} An X-ray imaging system is usually modeled as a pinhole camera~\cite{international2008radiotherapy,hartley2003multiple}, as illustrated in Figure \ref{fig:camera_model}, where the X-ray source serves as the camera center and the X-ray detector serves as the image plane. Following the convention in X-ray imaging~\cite{international2008radiotherapy}, we assume an isocenter coordinate system whose origin lies at the isocenter. Without loss of generality, we also assume the imaging model is calibrated, and there is no X-ray source offset and detector offset. Thus, the X-ray source, the isocenter, and the detector's origin are collinear, and the line from the X-ray source to the isocenter (referred to as the principal axis) is perpendicular to the detector. Let $d$ denote the distance between the X-ray source and the detector origin, and $c$ denote the distance between the X-ray source and the isocenter, then for a point $\mathbf{X} = (X, Y, Z)^T$ in the isocenter coordinate, its projection $\mathbf{x}$ on the detector is given by
\begin{equation} \label{eq:proj}
    \mathbf{x}' = \mathbf{K}
    \begin{bmatrix}
    \mathbf{I} & \mathbf{h}
    \end{bmatrix}
    \begin{pmatrix} \mathbf{X} \\ 1 \end{pmatrix},
\end{equation}
where
\begin{equation*}
\mathbf{K} = \begin{bmatrix}
    -d & 0 & 0\\
    0 & -d & 0\\
    0 & 0 & 1
\end{bmatrix}, \mathbf{h} = \begin{pmatrix}
    0 \\ 0 \\ -c
\end{pmatrix}.
\end{equation*}
Here $\mathbf{x}'= (x', y', z')$ is defined under the homogeneous coordinate and its counterpart under the detector coordinate can be written as $\mathbf{x} = (x, y) = (x'/z', y'/z')$.

In general, an X-ray is usually not captured at the canonical view as discussed above. Let $\mathbf{T}_{\text{view}}$ be a transformation matrix that converts a canonical view to a non-canonical view (Figure \ref{fig:camera_model}), then the projection of $\mathbf{X}$ for the non-canonical view can be written as
\begin{equation} \label{eq:proj_trans}
    \mathbf{x}' = \mathbf{K}
    \begin{bmatrix}
    \mathbf{R}_{\text{view}} & \mathbf{t}_{\text{view}} + \mathbf{h}
    \end{bmatrix}
    \begin{pmatrix} \mathbf{X} \\ 1 \end{pmatrix},
\end{equation}
where $\mathbf{R}_{\text{view}}$ and $\mathbf{\mathbf{t}_{\text{view}}}$ perform the rotation and translation, respectively, as in Equation (\ref{eq:trans}). Similarly, we can rewrite Equation (\ref{eq:drr}) at a non-canonical view as
\begin{equation} \label{eq:drr_view}
    \mathbf{I}_{\text{view}}^{\text{D}}(\mathbf{x}) = \int_{\mathbf{p} \in \bm{l}(\mathbf{x})} \mathbf{V}(\mathbf{T}^{-1} \mathbf{T}_{\text{view}}^{-1} \mathbf{p}) d\mathbf{p}.
\end{equation}

\subsection{The Proposed POINT$^2$ Approach}

An overview of the proposed method with two views is shown in Figure \ref{fig:overview}. Given a set of DRR and X-ray pairs of different views, our approach first selects a set of POIs in 3D from the CT volume and projects them to each DRR using Equation (\ref{eq:proj_trans}) as shown in Figure \ref{fig:overview}(a). Then, the approach measures the misalignment between each pair of DRR and X-ray by tracking the projected DRR POIs from the X-ray (Figure \ref{fig:overview}(b)). Using the tracked POIs on the X-rays, we can estimate their corresponding 3D POIs on the patient through triangulation (Figure \ref{fig:overview}(c)). Finally, by aligning CT POIs with patient POIs, the pose misalignment $\mathbf{T}^*$ between the CT and the patient can be calculated (Figure \ref{fig:overview}(d)).

{\vspace{0.5em} \noindent \bf POINT.} One of the key components of the proposed method is a Point-Of-Interest Network for Tracking (POINT) that finds the point-to-point correspondence between two images, that is, we use this network to track the POIs from DRR to X-ray. Specifically, the network takes a DRR and X-ray pair $(\mathbf{I}^{\text{D}}, \mathbf{I}^{\text{X}})$ and a set of projected DRR POIs $\{\mathbf{x}^{\text{D}}_1, \mathbf{x}^{\text{D}}_2, \dots, \mathbf{x}^{\text{D}}_m\}$ as the input and outputs the tracked X-ray POIs in the form of heatmaps $\{\hat{\mathbf{M}}^{\text{X}}_1, \hat{\mathbf{M}}^{\text{X}}_2, \dots, \hat{\mathbf{M}}^{\text{X}}_m\}$.

The structure of the network is illustrated in Figure \ref{fig:framework_2d_tracking}. We construct this network under a Siamese architecture~\cite{bertinetto2016fully,valmadre2017end} with each branch $\phi$ having an U-Net like structure~\cite{ronneberger2015u}. The weights of the two branches are shared. Each branch takes an image as the input and performs fine-grained feature extraction at pixel-level. Thus, the output is a feature map with the same resolution as the input image, and for an image with size M$\times$N, the size of the feature map is M$\times$N$\times$C where C is the number of channels. We denote the extracted feature maps of DRR and X-ray as $\mathbf{F}^{\text{D}} = \phi(\mathbf{I}^{\text{D}})$ and $\mathbf{F}^{\text{X}} = \phi(\mathbf{I}^{\text{X}})$, respectively.

With feature map $\mathbf{F}^{\text{D}}$, the feature vector of a DRR POI $\mathbf{x}^{\text{D}}_i$ can be extracted by interpolating $\mathbf{F}^{\text{D}}$ at $\mathbf{x}^{\text{D}}_i$. The feature extraction layer (FE layer) in Figure \ref{fig:framework_2d_tracking} performs this operation and we denote its output as a feature kernel $\mathbf{F}^{\text{D}}(\mathbf{x}^{\text{D}}_i)$. For a richer feature representation, the neighbor feature vectors around $\mathbf{x}^{\text{D}}_i$ may also be used. A neighbor of size $K$ gives in total (2K+1)$\times$(2K+1) feature vectors and the feature kernel $\mathbf{F}^{\text{D}}(\mathbf{x}^{\text{D}}_i)$ in this case has a size (2K+1)$\times$(2K+1)$\times$C.

Similarly, a feature kernel at $\mathbf{x}$ of the X-ray feature map can be extracted and denoted as $\mathbf{F}^{\text{X}}(\mathbf{x})$. Then, we may apply a similarity operation to $\mathbf{F}^{\text{D}}(\mathbf{x}^{\text{D}}_i)$ and $\mathbf{F}^{\text{X}}(\mathbf{x})$ to give a similarity score of the two locations $\mathbf{x}^{\text{D}}_i$ and $\mathbf{x}$. When the similarity check is operated exhaustively over all locations on the X-ray, the location $\mathbf{x}^*$ with the highest similarity score is regarded as the corresponding POI of $\mathbf{x}^{\text{D}}_i$ on the X-ray. Such an exhaustive search on $\mathbf{F}^{\text{X}}$ can be performed effectively with convolution and is denoted as a POI convolution layer in Figure \ref{fig:framework_2d_tracking}. The output of the layer is a heatmap $\hat{\mathbf{M}}^{\text{X}}_i$ and is computed by
\begin{equation} \label{eq:poi_conv}
    \hat{\mathbf{M}}^{\text{X}}_i = \mathbf{F}^{\text{X}} * (\mathbf{W} \odot \mathbf{F}^{\text{D}}(\mathbf{x}^{\text{D}}_i)),
\end{equation}
where $\mathbf{W}$ is a learned weight that selects the features for better similarity. Each element $\hat{\mathbf{M}}^{\text{X}}_i(\mathbf{x})$ denotes a similarity score of the corresponding location $\mathbf{x}$ on the X-ray.
\begin{figure}[t!]
  \centering
  \includegraphics[width=\linewidth]{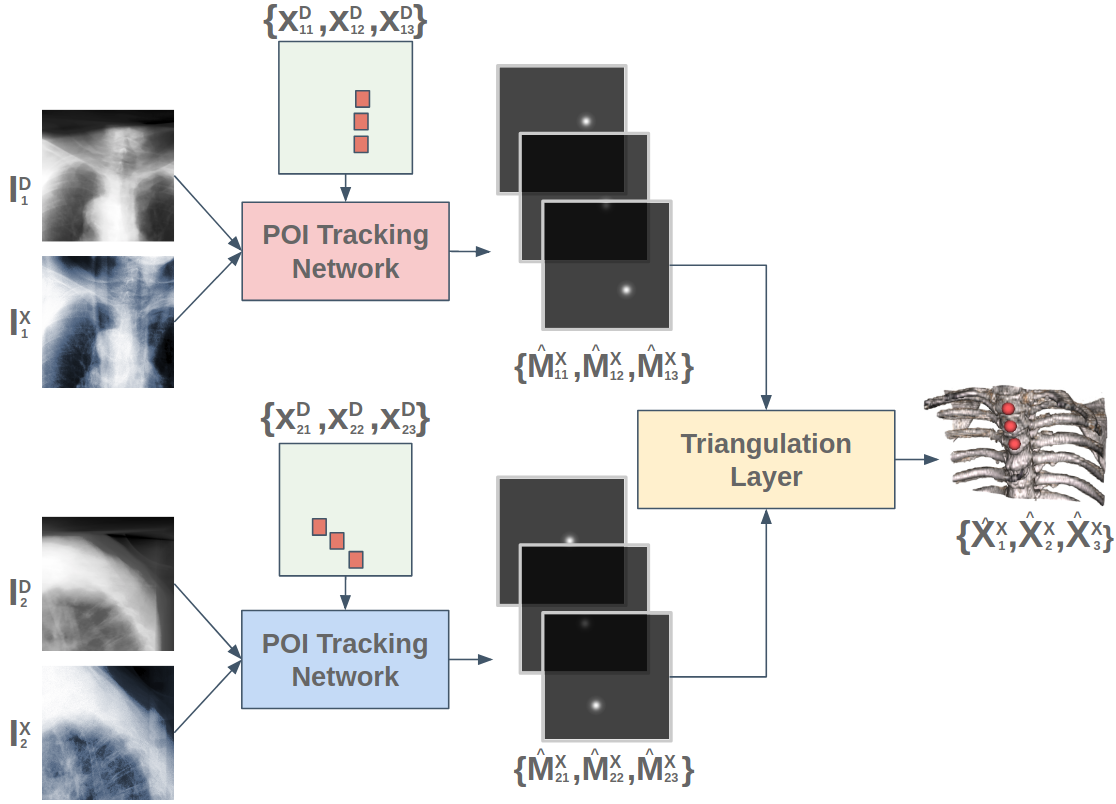}
  \caption{The overall framework of  POINT$^2$.}
\label{fig:framework_3d_tracking}
\end{figure}
{\vspace{0.5em} \noindent \bf POINT$^2$.} With the tracked POIs from different views of X-rays, we can obtain their 3D locations on the patient using triangulation as shown in Figure \ref{fig:overview}(c). However, this work seeks a uniform solution that formulates the POINT network and the triangulation under the same framework so that the two tasks can be trained jointly in an end-to-end fashion which could potentially benefit the learning of the tracking network. An illustration of this end-to-end design for two views is shown in Figure \ref{fig:framework_3d_tracking}. For an $n$-view 2D/3D registration problem, the proposed design will include $n$ POINT networks as discussed above. Each of the networks will track POIs for the designated view and, therefore, the weights are not shared among the networks. Given a set of DRR and X-ray pairs $\{(\mathbf{I}_{1}^{\text{D}}, \mathbf{I}_{1}^{\text{X}}), (\mathbf{I}_{2}^{\text{D}}, \mathbf{I}_{2}^{\text{X}}), \dots, (\mathbf{I}_{n}^{\text{D}}, \mathbf{I}_{n}^{\text{X}})\}$ of the $n$ views, these networks output the tracked X-ray POIs of each view in the form of heatmaps.

After obtaining the heatmaps, we introduce a triangulation layer that localizes a 3D point by forming triangles to it from the 2D tracked POIs from the heatmaps. Formally, we denote $\mathcal{M}_{j} = $ $\{\hat{\mathbf{M}}_{1j}^{\text{X}},$ $ \hat{\mathbf{M}}_{2j}^{\text{X}}, $ $ \dots, \hat{\mathbf{M}}_{nj}^{\text{X}}\}$ the set of heatmaps from different views but all corresponding to the same 3D POI $\hat{\mathbf{X}}_j^{\text{X}}$. Here, $\hat{\mathbf{M}}_{ij}^{\text{X}}$ is the heatmap of the $j$-th X-ray POI from the $i$-th view, and we obtain the 2D X-ray POI by
\begin{equation} \label{eq:map2poi}
    \hat{\mathbf{x}}^{\text{X}}_{ij} = \frac{1}{\sum_{\mathbf{x}}\hat{\mathbf{M}}_{ij}^{\text{X}}(\mathbf{x})} \sum_{\mathbf{x}}\hat{\mathbf{M}}_{ij}^{\text{X}}(\mathbf{x})\mathbf{x}.
\end{equation}
Next, we rewrite Equation (\ref{eq:proj_trans}) as
\begin{equation}
\mathbf{D}(\mathbf{x})\mathbf{R}_{\text{view}}\mathbf{X} = c\mathbf{x} - \mathbf{D}(\mathbf{x})\mathbf{t}_{\text{view}}, \label{eq:proj_trans2}
\end{equation} 
where
\begin{equation*}
\mathbf{D}(\mathbf{x}) =
\begin{bmatrix}
\begin{matrix}
d & 0 \\
0 & d
\end{matrix}
&
\mathbf{x}
\end{bmatrix}.
\end{equation*}
Thus, by applying Equation (\ref{eq:proj_trans2})  for each view, we can get
\begin{equation}
\begin{cases}
\mathbf{D}(\hat{\mathbf{x}}^{\text{X}}_{1j})\mathbf{R}_{1}\hat{\mathbf{X}}_j^{\text{X}} &= c\hat{\mathbf{x}}^{\text{X}}_{1j} - \mathbf{D}(\hat{\mathbf{x}}^{\text{X}}_{1j})\mathbf{t}_{1},\\
\mathbf{D}(\hat{\mathbf{x}}^{\text{X}}_{2j})\mathbf{R}_{2}\hat{\mathbf{X}}_j^{\text{X}} &= c\hat{\mathbf{x}}^{\text{X}}_{2j} - \mathbf{D}(\hat{\mathbf{x}}^{\text{X}}_{2j})\mathbf{t}_{2},\\
&\vdots \\
\mathbf{D}(\hat{\mathbf{x}}^{\text{X}}_{nj})\mathbf{R}_{n}\hat{\mathbf{X}}_j^{\text{X}} &= c\hat{\mathbf{x}}^{\text{X}}_{nj} - \mathbf{D}(\hat{\mathbf{x}}^{\text{X}}_{nj})\mathbf{t}_{n}.
\end{cases}
\end{equation}
Let
\begin{equation}
    \mathbf{A} = \begin{bmatrix}
        \mathbf{D}(\hat{\mathbf{x}}^{\text{X}}_{1j})\mathbf{R}_{1}\\
        \mathbf{D}(\hat{\mathbf{x}}^{\text{X}}_{2j})\mathbf{R}_{2}\\
        \vdots\\
        \mathbf{D}(\hat{\mathbf{x}}^{\text{X}}_{nj})\mathbf{R}_{n}
    \end{bmatrix},
    \mathbf{b} =
    \begin{bmatrix}
    c\hat{\mathbf{x}}^{\text{X}}_{1j} - \mathbf{D}(\hat{\mathbf{x}}^{\text{X}}_{1j})\mathbf{t}_{1} \\
    c\hat{\mathbf{x}}^{\text{X}}_{2j} - \mathbf{D}(\hat{\mathbf{x}}^{\text{X}}_{2j})\mathbf{t}_{2} \\
    \vdots \\
    c\hat{\mathbf{x}}^{\text{X}}_{nj} - \mathbf{D}(\hat{\mathbf{x}}^{\text{X}}_{nj})\mathbf{t}_{n}
    \end{bmatrix},
\end{equation}
then $\hat{\mathbf{X}}_j^{\text{X}}$ is given by
\begin{equation} \label{eq:backproject}
\hat{\mathbf{X}}_j^{\text{X}} = \mathbf{A}^{+}\mathbf{b}.
\end{equation}

The triangulation can be plugged into a loss function that regulates the training of POINT networks of different views. \begin{multline}
    \mathcal{L} = \frac{1}{mn}\sum_i\sum_j\text{BCE}(\sigma(\hat{\mathbf{M}}_{ij}^{\text{X}}), \sigma(\mathbf{M}_{ij}^{\text{X}})) \\ + \frac{w}{n}\sum_j ||\hat{\mathbf{X}}_j^{\text{X}} - \mathbf{X}_j^{\text{X}}||_2,
\end{multline}
where $\mathbf{M}_{ij}^{\text{X}}$ is the ground truth heatmap, $\mathbf{X}_j^{\text{X}}$ is the ground truth 3D POI, $\text{BCE}$ is the pixel-wise binary cross entropy function, $\sigma$ is the sigmoid function, and $w$ is a weight balancing the losses between tracking and triangulation errors.

{\vspace{0.5em} \noindent \bf Shape Alignment.} Let $\mathbf{P}^{\text{D}} = [\mathbf{X}_1^{\text{D}}~ \mathbf{X}_2^{\text{D}}~\dots~\mathbf{X}_m^{\text{D}}]$ be the selected CT POIs and $\mathbf{P}^{\text{X}} = [\hat{\mathbf{X}}_1^{\text{X}}~\hat{\mathbf{X}}_2^{\text{X}}~\dots~ \hat{\mathbf{X}}_m^{\text{X}}]$ be the estimated 3D POIs \footnote{The shape alignment assumes the points are under the homogeneous coordinate.}. The shape alignment finds a transformation matrix $\mathbf{T}^*$ such that the transformed $\mathbf{P}^{\text{D}}$ aligns closely with $\mathbf{P}^{\text{X}}$, i.e.,
\begin{equation}
     \mathbf{T}^* = \arg \min_{\mathbf{T}} ||\mathbf{T}\mathbf{P}^{\text{D}} - \mathbf{P}^{\text{X}}||_F, \text{ s.t., } \mathbf{R}\mathbf{R}^T = \mathbf{I}
\end{equation}
This problem is solved analytically through Procrustes analysis~\cite{seber2009multivariate}.

\section{Experiments}

\subsection{Dataset}

The dataset we use in the experiments is a cone-beam CT (CBCT) dataset captured for radiation therapy. The dataset contains 340 raw CBCT scans with each has 780 X-ray images. Each X-ray image comes with a geometry file that provides the registration ground truth as well as the information to reconstruct the CBCT volume. Each CBCT volume is reconstructed from the 780 X-ray images, and in total, we have 340 CBCT volumes (one for each CBCT scan). We use 300 scans for training and validation, and 40 scans for testing.  The size of the CBCT volumes is $448 \times 448 \times 768$ with 0.5 mm voxel spacing, and the size of the X-ray images is $512 \times 512$ with 0.388 mm pixel spacing. During the experiments, the CBCT volumes are treated as the 3D pre-intervention data, and the corresponding X-ray images are treated as the 2D intra-intervention data. Sample X-ray images from our dataset are shown in Figure. Note that unlike many existing approaches~\cite{otake2012intraoperative,pernus20133d,wang2017dynamic} that evaluate their methods on small datasets (typically about 10 scans) which are captured under relatively ideal scenarios, we use a significantly larger dataset with complex clinical settings, e.g., diverse field-of-views, surgical instruments/implants, various image contrast and quality, etc.

We consider two common views during the experiment: the anterior-posterior view and the lateral view. Hence, only X-rays that are close to ($\pm5\degree$) these views are used for training and testing. Note that this selection does not tightly constrain the diversity of the X-rays as the patient may be subject to movements with regard to the operating bed. To train the proposed method, X-ray and DRR pairs are selected and generated with a maximum of $10\degree$ rotation offset and $20$ mm translation offset. We first invert all the raw X-ray images and then apply histogram equalization to both the inverted X-ray images and DRRs to facilitate the similarity measurement. For each of the scan, we also annotate their landmarks on the reconstructed CBCT volume for further evaluation.

\begin{figure}[t!]
\centering
\begin{subfigure}{0.24\linewidth}
  \centering
  \includegraphics[width=\linewidth]{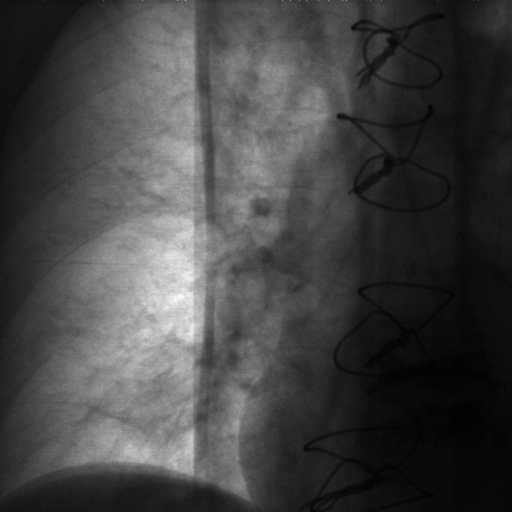}
\end{subfigure}
\begin{subfigure}{0.24\linewidth}
  \centering
  \includegraphics[width=\linewidth]{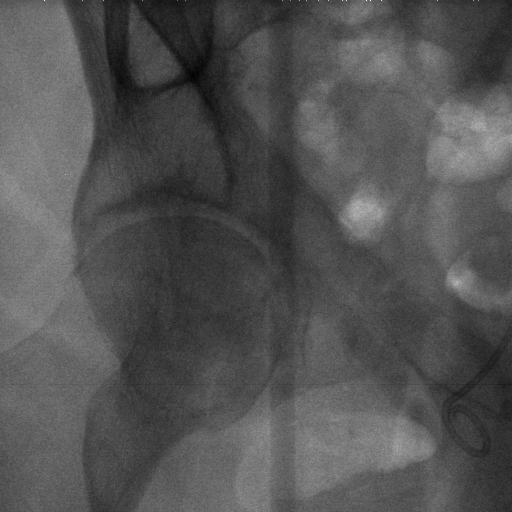}
\end{subfigure}
\begin{subfigure}{0.24\linewidth}
  \centering
  \includegraphics[width=\linewidth]{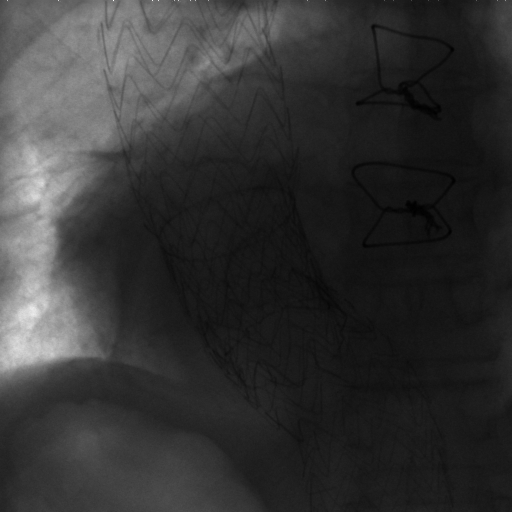}
\end{subfigure}
\begin{subfigure}{0.24\linewidth}
  \centering
  \includegraphics[width=\linewidth]{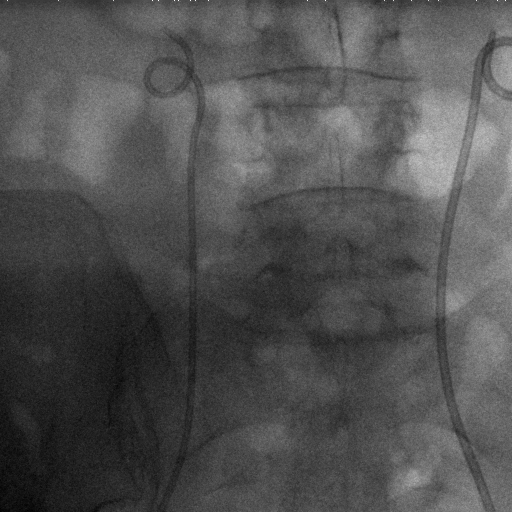}
\end{subfigure}
\caption{Sample raw X-ray images of our dataset.}
\label{fig:raw}
\end{figure}

\subsection{Implementation and Training Details}

We implement the proposed approach under the Pytorch\footnote{https://pytorch.org} framework with GPU acceleration. For the POINT network, each of the Siamese branch $\phi$ has five encoding blocks (BatchNorm, Conv, and LeakyReLU) followed by five decoding blocks (BatchNorm, Deconv, and ReLU), thus forming a symmetric structure, and we use skip-connections to shuttle the lower-level features from an encoding block to its symmetric decoding counterpart (see details in the supplementary material). The triangulation layer is implemented according to Equation (\ref{eq:backproject}) with the backpropagation automatically supported by Pytorch. We train the proposed approach in a two-stage fashion. In the first stage, we train the POINT network of each view independently for 30 epochs. Then, we fine-tune POINT$^2$ for 20 epochs. We find this mechanism converges faster than training POINT$^2$ from scratch. For the optimization, we use the mini-batch stochastic gradient descent with $0.01$ learning rate for the first stage and $0.001$ for the second. We set the loss weight as $w = 0.01$, which we empirically find it works well during training. For the X-ray imaging model, we use $d=1,500$ mm and $c=1,000$ mm.

\begin{figure}[t!]
\centering
\begin{subfigure}{0.33\linewidth}
  \centering
  \includegraphics[width=1\linewidth]{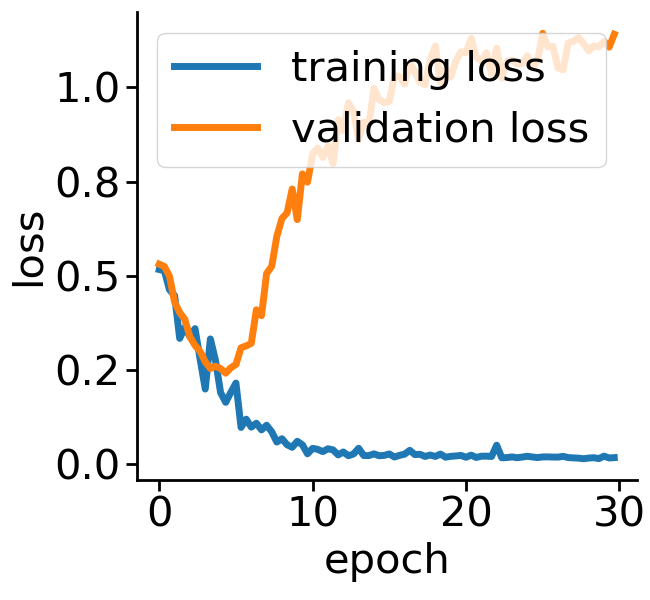}
  \caption{landmark}
\end{subfigure}%
\begin{subfigure}{0.33\linewidth}
  \centering
  \includegraphics[width=1\linewidth]{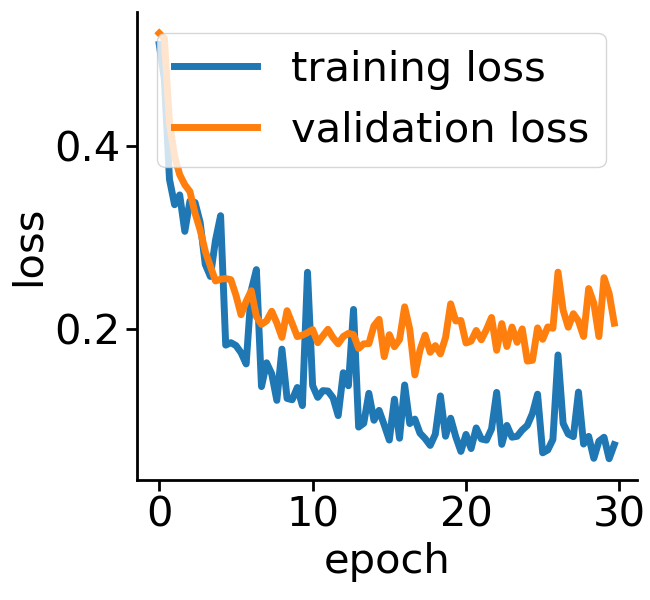}
  \caption{Harris corner}
\end{subfigure}
\begin{subfigure}{0.33\linewidth}
  \centering
  \includegraphics[width=1\linewidth]{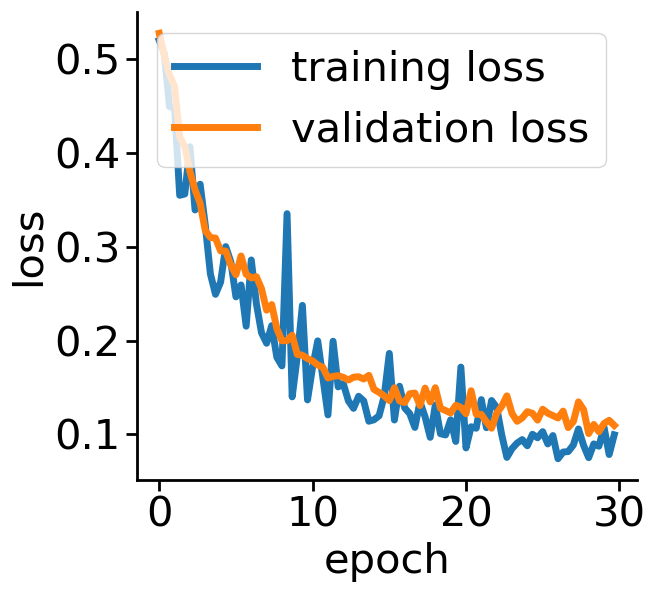}
  \caption{random}
\end{subfigure}
\caption{Training and validation losses of different POI selection methods.}
\label{fig:poi_selection}
\end{figure}

\subsection{Ablation Study}

This section discusses an ablation study of the proposed POINT network. As the network tracks POIs in 2D, we use mean projected distance (mPD)~\cite{van2005standardized} to evaluate different models with specific design choices. The evaluation results are given in Table \ref{tab:ablation_study}.

{\vspace{0.5em} \noindent \bf POI Selection.} The first step of the proposed approach requires selecting a set of POIs to set up a point-to-point correspondence. In this experiment, we investigate different POI selection strategies. First, we investigate directly using landmarks as the POIs since they usually have strong semantic meaning and can be annotated before the intervention. Second, we also investigate an automatic solution that uses the Harris corners as the POIs to avoid the labor work of annotation. Finally, we try random POI selection.

As shown in Figure \ref{fig:poi_selection} (a), we find our approach is prone to overfitting when trained with landmark POIs. This is actually reasonable as each CBCT volume only contains about a dozen of landmarks, which in total is about $3,000$ POIs. Considering the variety of the field of views of our dataset, this is far from enough and leads to the overfitting. For the Harris corners, a few hundreds of POIs are selected from each CBCT volume, and we can see an improvement in performance, but the overfitting still exists (Figure \ref{fig:poi_selection} (b)). We find the use of random POIs gives the best performance and generalizes well to unseen data (Figure \ref{fig:poi_selection} (c)). This seemly surprising observation is, in fact, reasonable as it forces the model to learn a more general way to extract features at a fine-grained level, instead of memorizing some feature points that may look different when projected from a different view.

{\vspace{0.5em} \noindent \bf POI Convolution.} We also explore two design options for the POI convolution layer. First, it is worth knowing that how much neighborhood information around the POI is necessary to extract a distinctive feature while the learning can still be easily generalized. To this end, we try different sizes of the feature kernel for POI convolution as given in Equation (\ref{eq:poi_conv}). Rows 1-3 in Table \ref{tab:ablation_study} show the performance of the POINT network with different feature kernel sizes. We observe that a $1\times1$ kernel does not give features distinctive enough for better similarity measure and a  $5\times5$ kernel seems to include too much neighborhood information (and use more computation) that is harder for the model to figure out a general representation. In general, a $3\times3$ kernel serves better for the feature similarity measure. It should also be noted that a $1\times1$ kernel does not mean only the information at the current pixel location is used since each element of $\mathbf{F}^{\text{D}}$ or $\mathbf{F}^{\text{X}}$ is supported by the receptive field of the U-Net that readily provides rich neighborhood information. Second, we compare the performance of the POINT network with or without having the weight $W$ in Equation (\ref{eq:poi_conv}). Rows 2 and 6 show that it is critical to have a weighted feature kernel convolution so that discriminate features can be highlighted in the similarity measure.

\begin{figure}[t!]
\centering
\begin{subfigure}{0.24\linewidth}
  \centering
  \includegraphics[width=1\linewidth]{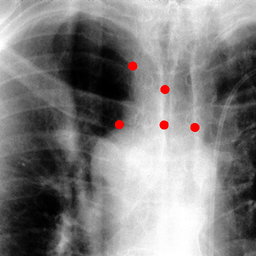}
  \captionsetup{labelformat=empty}
  \caption{}
\end{subfigure}
\begin{subfigure}{0.24\linewidth}
  \centering
  \includegraphics[width=1\linewidth]{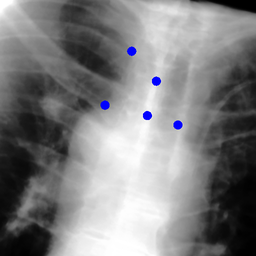}
  \captionsetup{labelformat=empty}
  \caption{13.6 $\rightarrow$ 7.3}
\end{subfigure}
\begin{subfigure}{0.24\linewidth}
  \centering
  \includegraphics[width=1\linewidth]{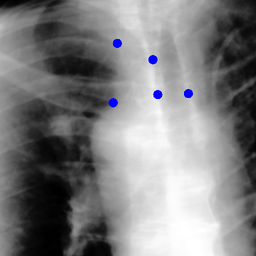}
  \captionsetup{labelformat=empty}
  \caption{22.9 $\rightarrow$ 9.0}
\end{subfigure}
\begin{subfigure}{0.24\linewidth}
  \centering
  \includegraphics[width=1\linewidth]{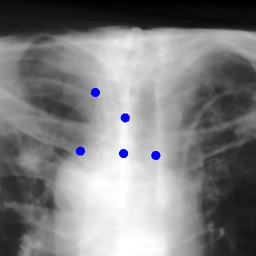}
  \captionsetup{labelformat=empty}
  \caption{37.1 $\rightarrow$ 7.8}
\end{subfigure}

\begin{subfigure}{0.24\linewidth}
  \centering
  \includegraphics[width=1\linewidth]{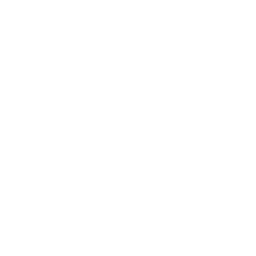}
\end{subfigure}
\begin{subfigure}{0.24\linewidth}
  \centering
  \includegraphics[width=1\linewidth]{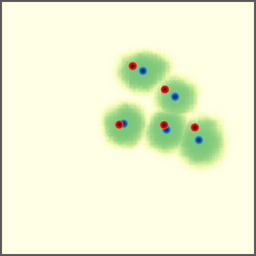}
\end{subfigure}
\begin{subfigure}{0.24\linewidth}
  \centering
  \includegraphics[width=1\linewidth]{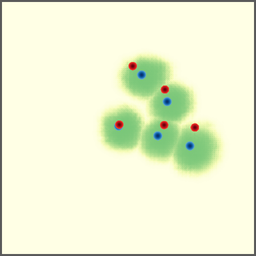}
\end{subfigure}
\begin{subfigure}{0.24\linewidth}
  \centering
  \includegraphics[width=1\linewidth]{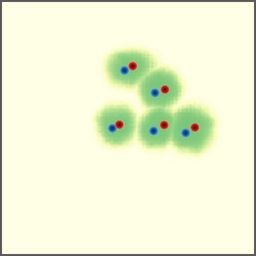}
\end{subfigure}

\begin{subfigure}{0.24\linewidth}
  \centering
  \includegraphics[width=1\linewidth]{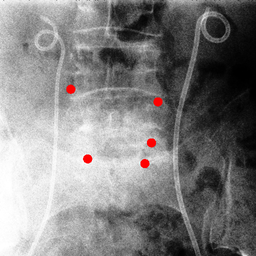}
  \captionsetup{labelformat=empty}
  \caption{}
\end{subfigure}
\begin{subfigure}{0.24\linewidth}
  \centering
  \includegraphics[width=1\linewidth]{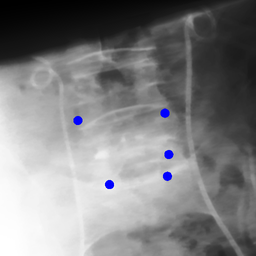}
  \captionsetup{labelformat=empty}
  \caption{19.5 $\rightarrow$ 8.7}
\end{subfigure}
\begin{subfigure}{0.24\linewidth}
  \centering
  \includegraphics[width=1\linewidth]{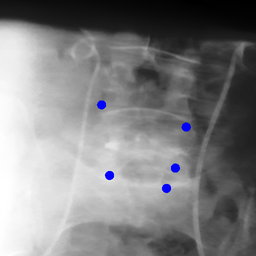}
  \captionsetup{labelformat=empty}
  \caption{26.0 $\rightarrow$ 9.5}
\end{subfigure}
\begin{subfigure}{0.24\linewidth}
  \centering
  \includegraphics[width=1\linewidth]{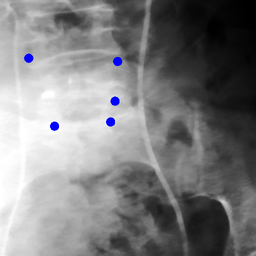}
  \captionsetup{labelformat=empty}
  \caption{41.1 $\rightarrow$ 11.4}
\end{subfigure}

\setcounter{subfigure}{0}
\begin{subfigure}{0.24\linewidth}
  \centering
  \includegraphics[width=1\linewidth]{white.png}
  \caption{}
\end{subfigure}
\begin{subfigure}{0.24\linewidth}
  \centering
  \includegraphics[width=1\linewidth]{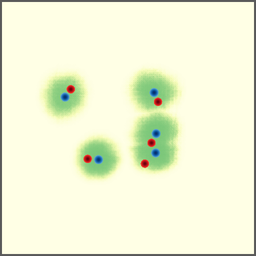}
  \caption{}
\end{subfigure}
\begin{subfigure}{0.24\linewidth}
  \centering
  \includegraphics[width=1\linewidth]{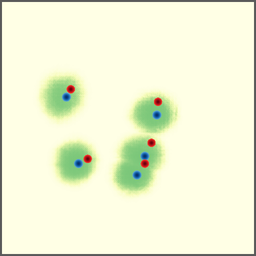}
  \caption{}
\end{subfigure}
\begin{subfigure}{0.24\linewidth}
  \centering
  \includegraphics[width=1\linewidth]{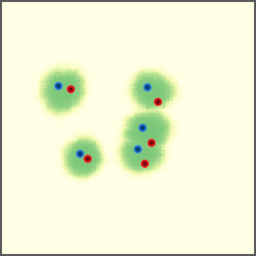}
  \caption{}
\end{subfigure}
\caption{POI tracking results. (a) X-ray image. (b-d) DRR images with different in-plane offsets. The heatmaps of the tracking results are all aligned with the X-ray images and appear similar, showing the shift-invariant property.}
\label{fig:shift_invariant}
\end{figure}

{\vspace{0.5em} \noindent \bf Shift-Invariant Tracking.} The POINT network benefits from the shift invariant property of the convolution operation, which makes it less sensitive to the in-plane offset of the DRRs. Figure \ref{fig:shift_invariant} shows some tracking results from the POINT network. Here the odd rows show the (a) X-ray and (b-d) DRR images. The heatmap below each DRR shows the tracking result between this DRR and the leftmost X-ray image. The red and the blue marks on the X-ray and DRR images denote the POIs. The red and the blue marks on the heatmaps are the ground truth POIs and the tracked POIs, respectively. The green blobs are the heatmap responses and they are used to generate the tracked POIs (blue) according to Equation (\ref{eq:map2poi}). The numbers under each DRR denote the mPD scores before and after the tracking. As we can observe that the tracking results are consistently good, no matter how much initial offset there is between the DRR and the X-ray image. This shows that our POINT network indeed benefits from the POI convolution layer and provide more consistent outputs regardless of the in-plane offsets.

\begin{table}[]
\centering
\caption{Ablation study of the proposed POINT network.}
\resizebox{0.98\columnwidth}{!}{%
\begin{tabular}{@{}c?ccc?ccc?cc?c@{}}
\noalign{\hrule height 1pt}
\multicolumn{1}{m{1em}?}{\vspace{0.7em} \# \vspace{-0.7em}} & \multicolumn{3}{c?}{Kernel size}     & \multicolumn{3}{c?}{POI type}        & \multicolumn{2}{c?}{Weight} & mPD \hspace{0.3em}           \\
      & 1          & 3          & 5          & land.      & Harris     & rand.           & w/              & w/o              & (mm) \hspace{0.3em}          \\ \noalign{\hrule height 1pt}
         1 & \checkmark &            &            &            &            & \checkmark      & \checkmark      &                 & 8.46          \\
         2 &            & \checkmark &            &            &            & \checkmark      & \checkmark      &                 & \textbf{8.12} \\
         3 &            &            & \checkmark &            &            & \checkmark      & \checkmark      &                 & 9.49          \\
         4 &            & \checkmark &            &            & \checkmark &                 & \checkmark      &                 & 9.87          \\
         5 &            & \checkmark &            & \checkmark &            &                 & \checkmark      &                 & 12.72         \\
         6 &            & \checkmark &            &            &            & \checkmark      &                 & \checkmark      & 11.26         \\ \noalign{\hrule height 1pt}
\end{tabular}%
}
\label{tab:ablation_study}
\end{table}

\subsection{2D/3D Registration}

We compare our method with one learning-based (MDP~\cite{DBLP:conf/aaai/MiaoPFTMML18}) and three optimization-based methods (Opt-GC~\cite{de20163d}, Opt-GO~\cite{de20163d} and Opt-NGI~\cite{otake2013robust}). To further evaluate the performances of the proposed method as an initial pose estimator, we also compare two approaches that use MDP or our method to initialize the optimization. We denote these two approaches as MDP+opt and POINT$^2$+opt, respectively. Finally, we investigate the registration performance of our method that only uses the POINT network without the triangulation layer, and denote the corresponding models as POINT and POINT+opt. For MDP+opt, POINT+opt and POINT$^2$+opt, we use the Opt-GC method during the optimization as we find it converges faster when the initial pose is close to the global optima.

Following the standard in 2D/3D registration~\cite{van2005standardized}, the performances of the proposed method and the baseline methods are evaluated with mean target registration error (mTRE), i.e., the mean distance (in mm) between the patient landmarks and the aligned CT landmarks in 3D. The mTRE results are reported in forms of the 50th, 75th, and 95th percentiles to demonstrate the robustness of the compared methods. In addition, we also report the gross failure rate (GFR) and average registration time, where GFR is defined as the percentage of the tested cases with a TRE greater than 10 mm~\cite{DBLP:conf/aaai/MiaoPFTMML18}.

The evaluation results are given in Table \ref{tab:registration}. We find that the optimization-based methods generally require a good initialization for accurate registration. Otherwise, they fail quickly. Opt-NGI overall is less sensitive to the initial location than Opt-GO and Opt-GC, with more than half of the registration results have less than 1 mm mTRE. Despite the high accuracy, it still suffers from the high failure rate and long registration time and so do the Opt-GO and Opt-GC methods. On the other hand, MDP achieves a better GFR and registration time by learning a function that guides the iterative pose searching. This also demonstrates the benefit of using a learning-based approach to guide the registration. However, due to the problems we have mentioned in Section \ref{sec:intro}, it still has a relatively high GFR and a noticeable registration time. In contrast, our base model POINT already achieves comparable performance to MDP; however, it runs over twice faster. Further, by including the triangulation layer, POINT$^2$ performs significantly better than both POINT and MDP in terms of mTRE and GFR. It means that the triangulation layer that brings the 3D information to the training of the POINT network is indeed useful.

In addition, we notice that when our method is combined with an optimization-based method (POINT$^2$ + Opt) the GFR is greatly reduced, which demonstrates that our method provides initial poses that are close to the global optima such that the optimization is unlikely to fall into local optima. The speed is also significantly improved due to faster convergence and less sampling over the pose space. 

\begin{table}[]
\centering
\caption{2D/3D registration performance comparing with the state-of-the-art results.}
\begin{tabular}{@{}c?ccccc@{}}
\noalign{\hrule height 1pt}
\multicolumn{1}{l?}{\multirow{2}{*}{}} & \multicolumn{3}{c}{mTRE (mm)} & \multirow{2}{*}{GFR} & Reg. \\
\multicolumn{1}{l?}{}                  & 50th   & 75th   & 95th   &                      & time      \\ \noalign{\hrule height 1pt}
Initial                                & 20.4   & 24.4   & 29.7   & 92.9\%               & N/A       \\
Opt-NGI~\cite{otake2013robust}                                   & \textbf{0.62}   & 25.2   & 57.8   & 40.0\%               & 23.5s     \\
Opt-GO~\cite{de20163d}                                    & 6.53   & 23.8   & 44.7   & 45.1\%               & 22.8s      \\
Opt-GC~\cite{de20163d}                                 & 7.40   & 25.7   & 56.5   & 47.7\%               & 22.1s      \\
MDP~\cite{DBLP:conf/aaai/MiaoPFTMML18} & 5.40   & 8.62   & 27.6   & \underline{16.4\%}               & 1.74s          \\
POINT                               & 5.63   & \underline{7.72}   & \underline{12.8}  & 18.6\%               & \textbf{0.75s}      \\
POINT$^2$                          & \underline{4.22}   & \textbf{5.70}   & \textbf{9.84}   & \textbf{4.9\%}                & \underline{0.78s}      \\ \noalign{\hrule height 1pt}
MDP~\cite{DBLP:conf/aaai/MiaoPFTMML18} + Opt                             & \underline{1.06}      & \underline{2.25}      & 24.6      & 15.6\%                     & 3.21s          \\
POINT + Opt                         & 1.19   & 4.67   & \underline{21.8}  & \underline{14.8}\%               & \textbf{2.16s}     \\
POINT$^2$ + Opt                    & \textbf{0.55}   & \textbf{0.96}   & \textbf{5.67}   & \textbf{2.7\%}                & \underline{2.25s}      \\ \noalign{\hrule height 1pt}
\end{tabular}
\label{tab:registration}
\end{table}

\section{Limitations}

First, similar to other learning-based approaches, our method requires a considerably large dataset from the targeting medical domain for learning reliable feature representations. When the data is insufficient, the proposed method may fail. Second, although our method alone is quite robust
and its accuracy is state-of-the-art through a combination with the optimization-based approach, it is still desirable to come up with a more elegant solution to solve the problem directly. 
Finally, due to the use of triangulation, our method requires X-rays from at least two views to be available. Hence, for the applications where only a single view is acceptable, our method will render an estimate of registration parameter with inherent ambiguity. 
    
\section{Conclusion}

We proposed a fast and robust method for 2D/3D registration. The proposed method avoids the often costly and unreliable iterative pose searching by directly aligning the CT with the patient through a novel POINT$^2$ framework, which first establishes the point-to-point correspondence between the pre- and intra-intervention data in both 2D and 3D, and then performs a shape alignment between the matched points to estimate the pose of the CT. We evaluated the proposed POINT$^2$ framework on a challenging and large-scale CBCT dataset and showed that 1) a robust POINT network should be trained with random POIs, 2) a good POI convolution layer should be convolved with weighted $3\times3$ feature kernel, and 3) the POINT network is not sensitive to in-plane offsets. We also demonstrated that the proposed POINT$^2$ framework is significantly more robust and faster than the state-of-the-art learning-based approach. When used as an initial pose estimator, we also showed that the POINT$^2$ framework can greatly improve the speed and robustness of the current optimization-based approach while attaining a higher registration accuracy. Finally, we discussed several limitations of the POINT$^2$ framework which we will address in our future work.

{\vspace{0.5em} \noindent \bf Acknowledgement:} This work was partially supported by NSF award \#17228477, and the Morris K. Udall Center of Excellence in Parkinson's Disease Research by NIH.

{\small
\bibliographystyle{ieee_fullname}
\bibliography{references}
}

\end{document}